\documentclass{article}

\usepackage[final]{corl_2026} 

\usepackage{booktabs}
\usepackage{multirow}
\usepackage{graphicx}
\usepackage{amsmath}
\usepackage{amsfonts}
\usepackage{mathrsfs}
\usepackage{wrapfig}
\usepackage{hyperref}
\usepackage{enumitem}
\usepackage{caption}
\usepackage{bbm}
\usepackage{xcolor}
\usepackage{marvosym}

\title{GeoHAT: Geometry-Adaptive Hybrid Action Transformer for Mobile Manipulation}

\author{
    Xiangyu Zhu$^{1^*}$, Renjun Wu$^{1^*}$, Luzhou Ge$^{1^*}$, Jinyan Liu$^{1}\protect\textsuperscript{\Letter}$, Xuesong Li$^{1}\protect\textsuperscript{\Letter}$ \\ [0.5em]
    $^1$School of Computer Science, Beijing Institute of Technology, China \\ [0.5em]
    $^*$Equal contribution, \protect\textsuperscript{\Letter} Corresponding author \\ [0.5em]
    \href{https://icr-lab.github.io/GeoHAT/}{https://icr-lab.github.io/GeoHAT/}
}

\begin{document}
\maketitle
\begin{abstract}
Whole-body mobile manipulation requires coordinating mobile base and manipulator under shifting viewpoints, posing challenges in geometric perception and action generation. 
Current policies either rely on 2D features or sparse 3D representations that lack dense spatial structure, and typically encode arm and base within one action vector that ignores their distinct control demands.   
Moreover, existing dense
fusion strategies risk corrupting pretrained representations under noisy depth while incurring heavy computational overhead.
We present \textbf{GeoHAT}, an end-to-end diffusion-based framework built on a simple principle: geometry should be injected only where reliable and attended to only where needed. 
GeoHAT employs a lightweight Fourier spatial encoder that maps dense per-pixel 3D coordinates into geometric tokens without an additional 3D vision backbone. 
These tokens are then selectively injected into vision foundation model features through per-token gated fusion modulated by depth validity, preserving the semantic prior while enriching spatial understanding. 
For action generation, a Hybrid Whole-Body Action Decoder decomposes arm and base into distinct subspaces and lets each action modality attend to its task-relevant visual context through sparse cross-attention, while causal temporal modeling captures intra-timestep coordination and inter-timestep dependencies.
Experiments on the ManiSkill-HAB simulation benchmark demonstrate that GeoHAT achieves a 79.2\% mean success rate, surpassing the strongest baseline by 23.6~pp. Furthermore, real-world experiments on diverse tasks also confirm consistent improvements over all baselines.
\end{abstract}


\section{Introduction}

Whole-body mobile manipulation is a fundamental capability for general-purpose robotic agents operating in unstructured environments~\citep{AC-DiT,MoManipVLA,WB-VIMA,AnchorVLA}. Compared with fixed-base manipulators, it requires coordinating a mobile base and a manipulator under continuously shifting viewpoints, imposing stringent requirements on both spatial perception and action coordination~\citep{AnchorVLA,DSPv2}. Recent imitation learning~\citep{ACT,DP,DP3,M2Diffuser} and vision-language-action models~\citep{RT2,OpenVLA,Octo} have yielded strong results on tabletop tasks, yet extending them to whole-body robots remains formidable~\citep{MoManipVLA,DSPv2} due to two tightly coupled challenges. On the \emph{perception} side, contact-rich interactions---grasping a handle, aligning with a drawer, or positioning the base for reachability---require fine-grained 3D spatial awareness that 2D features~\citep{ACT,DP} or sparse point clouds~\citep{WB-VIMA,DP3} cannot reliably provide. On the \emph{action} side, the arm and base exhibit fundamentally different dynamics, and minor base inaccuracies propagate through the kinematic chain to become amplified at the end-effector~\citep{AnchorVLA,DSPv2}, making their coordination a first-order concern.

\begin{figure}[t]
    \centering
    \includegraphics[width=1.0\linewidth]{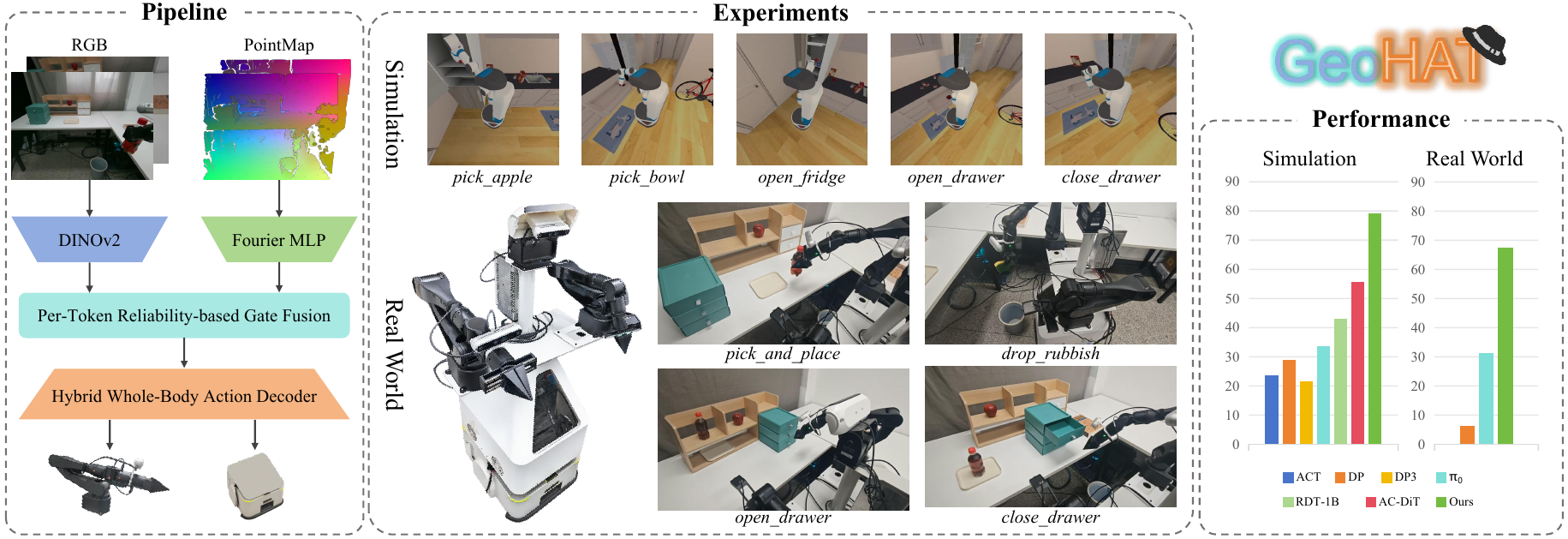}
    \vspace{-1em}
    \caption{\textbf{Overview of GeoHAT.}
    \textit{Left:} GeoHAT fuses multi-view RGB features with PointMap features through reliability-aware gated fusion, and then predicts coordinated arm-base actions with a hybrid whole-body decoder. This design injects geometry only where reliable and routes visual context to the action subspaces that need it. \textit{Middle:} Simulation and real-world mobile manipulation tasks used for evaluation. \textit{Right:} Success rates showing strong performance in both simulation and real-world deployment.}
    \label{fig:teaser}
    \vspace{-2.0em}
\end{figure}

To address the perception challenge, recent methods incorporate dense 3D representations into visuomotor policies~\citep{PointMapPolicy,GeoVLA,Lift3D,SpatialVLA}. However, two critical issues remain unsolved. First, prevailing fusion strategies---concatenation, addition, or fixed projection---treat all geometric signals uniformly regardless of depth quality~\citep{DSPv2,PointMapPolicy}. In mobile manipulation, where viewpoints shift continuously and depth sensors frequently produce invalid readings at reflective surfaces or object boundaries, such indiscriminate fusion risks injecting spurious geometry into pretrained visual representations. Heavy 3D encoders~\citep{PerAct,RVT} can improve spatial reasoning but introduce computational overhead incompatible with real-time control. Second, on the action side, end-to-end methods~\citep{AC-DiT,MoManipVLA,WB-VIMA,AnchorVLA} jointly optimize base and arm but most model both as a homogeneous action vector, while modular frameworks~\citep{SAGA,FALCON} decouple them at the cost of coordination complexity. Critically, existing decoders route identical visual features to all action dimensions---yet arm actions demand local contact geometry for precise grasping, while base actions require broader spatial context for positioning and navigation. This mismatch between uniform visual conditioning and heterogeneous action demands limits whole-body coordination.

These limitations point to a design principle: \emph{geometry should be injected only where it is reliable, and attended to only where it is needed}. We present \textbf{GeoHAT} (Geometry-Adaptive Hybrid Action Transformer), a diffusion-based imitation learning framework that instantiates this principle through three mechanisms. First, a \emph{lightweight Fourier spatial encoder} maps patch-aligned PointMap coordinates into geometric tokens without an additional vision backbone. Second, a \emph{reliability-aware gated fusion} injects geometric features into pretrained DINOv2 \citep{dinov2} tokens through per-token gating modulated by depth confidence, suppressing unreliable regions while preserving the semantic prior. Third, a \emph{Hybrid Whole-Body Action Decoder} decomposes the action space into arm and base subspaces and employs query-level sparse cross-attention, enabling each subspace to selectively attend to its task-relevant visual context. Causal masked self-attention further permits intra-timestep arm-base coordination while enforcing temporal causality. GeoHAT achieves a 79.2\% mean success rate on ManiSkill-HAB \cite{maniskill}, surpassing the strongest baseline by 23.6 percentage points (pp), with about $2\times$ improvement on multi-step tasks requiring base repositioning before manipulation.

In summary, our contributions are as follows:
\begin{itemize}[leftmargin=*, itemsep=2pt, topsep=0pt, parsep=0pt]
\item We propose GeoHAT, a diffusion-based imitation learning framework that enriches pretrained visual features with dense 3D geometry through a lightweight Fourier spatial encoder and per-token reliability-aware gated fusion, achieving strong spatial reasoning without heavy 3D encoders.
\item We design a Hybrid Whole-Body Action Decoder that decomposes arm and base into distinct subspaces, employs query-level sparse attention for subspace-specific visual grounding, and uses causal masked self-attention for intra-timestep coordination and inter-timestep consistency.
\item Extensive experiments on ManiSkill-HAB and real-world deployment demonstrate strong performance over representative baselines, with controlled ablations validating the synergistic contribution of each design component.
\end{itemize}

\section{Related Work}
\label{sec:related_work}

\textbf{3D Representations for Visuomotor Policies.}
Incorporating 3D information into visuomotor policies enhances geometric reasoning but introduces tradeoffs in cost and robustness. Voxel-based methods~\citep{PerAct} and multi-view projections~\citep{RVT,RVT2} are computationally expensive and target fixed-base keypose prediction. DP3~\citep{DP3} encodes sparse point clouds compactly but discards local structure; 3D Diffuser Actor~\citep{3DDiffuserActor} and Act3D~\citep{Act3D} tokenize scene points but remain limited to static viewpoints. For integrating 3D cues with pretrained 2D features, Lift3D~\citep{Lift3D} uses auxiliary 3D pretraining, PointVLA~\citep{PointVLA} and GeoVLA~\citep{GeoVLA} employ separate encoders with late fusion, SpatialVLA~\citep{SpatialVLA} injects spatial positional encodings, and DSPv2~\citep{DSPv2} aligns sparse 3D features via a Q-former. PointMapPolicy~\citep{PointMapPolicy} shows that dense per-pixel 3D maps benefit policy learning but encodes them with a ViT from scratch, adding substantial overhead. A common limitation across these approaches is that they rely on fixed fusion strategies---concatenation, addition, or learned projection---that treat all geometric signals uniformly regardless of depth quality. When depth observations are noisy or incomplete, as is common in mobile settings with shifting viewpoints, such uniform fusion can degrade the pretrained semantic representations rather than enhance them. GeoHAT addresses this through a lightweight Fourier encoder that avoids an additional backbone and per-token reliability-aware gating that selectively injects only trustworthy geometric signals, preserving the pretrained prior under unreliable depth.

\textbf{Whole-Body Mobile Manipulation Policies.}
Early systems decomposed tasks into navigation and manipulation phases via high-level planners~\citep{OKRobot,TAMP,SayCan,wu2025moto}, suffering from error accumulation when phases are tightly coupled. End-to-end methods jointly model base and arm: Mobile ALOHA~\citep{MobileALOHA} via whole-body teleoperation, diffusion policies~\citep{DP,DP3,M2Diffuser} via learned denoising, and VLA models~\citep{RT2,OpenVLA,Octo} via large-scale pretraining. For coordination specifically, AC-DiT~\citep{AC-DiT} introduces unidirectional mobility-to-body conditioning, WB-VIMA~\citep{WB-VIMA} adopts hierarchical autoregressive decoding, AnchorVLA~\citep{AnchorVLA} employs anchor-guided truncated diffusion but still models arm and base within a unified action space, and modular approaches~\citep{SAGA,MoManipVLA,FALCON} decouple subsystems via shared representations or trajectory optimization. Despite these advances, most methods either treat arm and base as a homogeneous vector or route identical visual features to all action dimensions regardless of what each subspace requires. GeoHAT decomposes the action space into heterogeneous subspaces with query-level sparse attention, letting each subspace attend to its task-relevant visual context while causal self-attention captures both intra-timestep coordination and inter-timestep consistency.

\section{Methodology}
\label{sec:method}

\noindent\textbf{Problem Formulation.}
    We formulate whole-body mobile manipulation as a partially observable Markov decision process. At timestep $t$, the multimodal observation $o_t = \{I_t, P_t, S_t\}$ comprises RGB images $I_t$ from head and wrist cameras, dense per-pixel 3D coordinate maps (PointMaps) $P_t$ in their respective camera frames, and the proprioceptive state $S_t$ consisting of joint positions, joint velocities, and the tool-center-point (TCP) pose. The action space is decomposed into two heterogeneous subspaces: $A^{\mathrm{arm}}$ for manipulation and $A^{\mathrm{base}}$ for locomotion. Our objective is to learn a policy $\pi(A_{t+1:t+h} \mid o_t)$ that predicts a coordinated whole-body action sequence over a prediction horizon $h$.
 
\begin{figure}[t]
    \centering
    \includegraphics[width=1.0\linewidth]{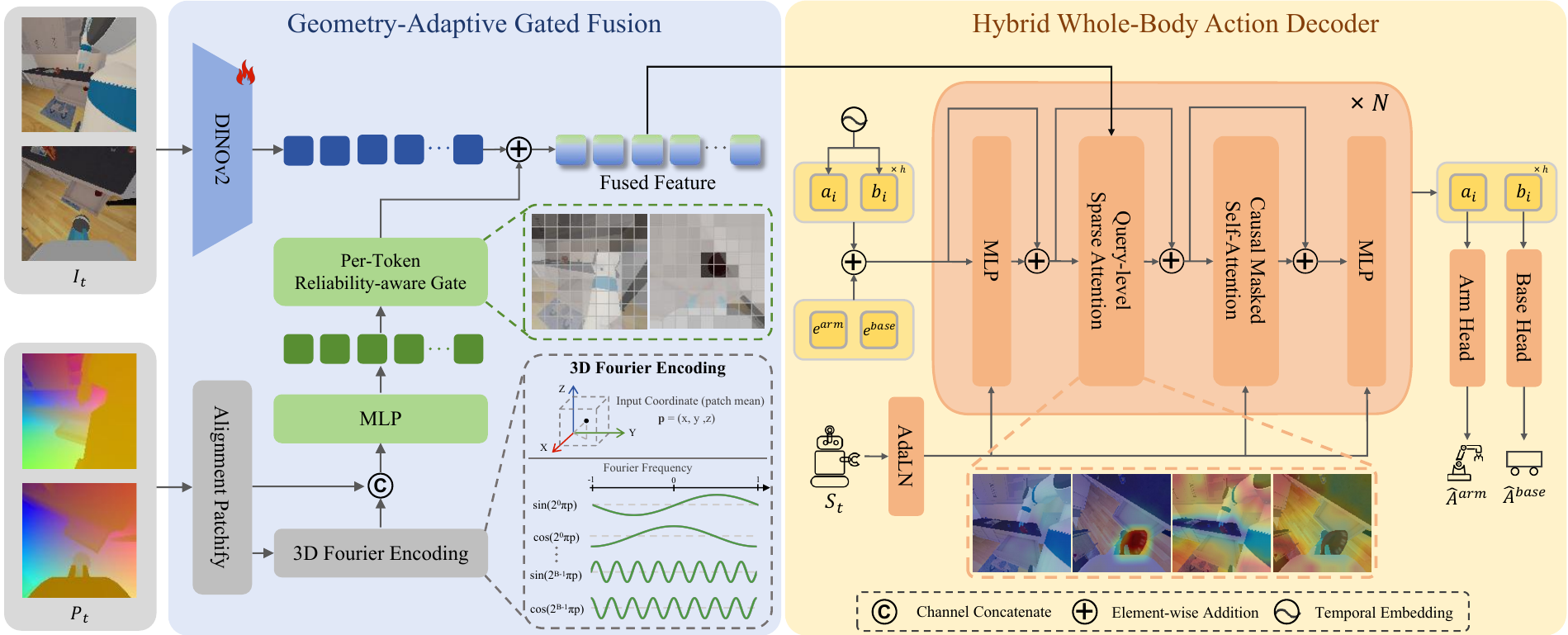}
    \vspace{-0.5em}
    \caption{\textbf{Architecture of GeoHAT.} Patch-aligned PointMap coordinates are encoded via a Fourier MLP and selectively fused into DINOv2 features through per-token gating modulated by depth confidence (\textit{left}). The hybrid action decoder interleaves arm and base tokens, applies query-level Top-$K$ cross-attention for subspace-specific visual grounding, and uses causal self-attention for temporal coordination (\textit{right}).}
    \label{fig:overview_geohat}
    \vspace{-1.0em}
\end{figure}

\noindent\textbf{Overview.}
    GeoHAT instantiates the principle that geometry should be injected only where it is reliable and attended to only where it is needed, as illustrated in Fig.~\ref{fig:overview_geohat}. In the first stage (Sec.~\ref{sec:encoder}), a DINOv2~\citep{dinov2} backbone extracts patch-level semantic features from each RGB view, while a lightweight Fourier spatial encoder maps the corresponding PointMap into patch-aligned geometric tokens. A per-token reliability-aware gate then selectively injects geometry into the semantic representation based on depth confidence. In the second stage (Sec.~\ref{sec:decoder}), a Hybrid Whole-Body Action Decoder decomposes the action space into arm and base subspaces, lets each subspace query its task-relevant visual context through sparse cross-attention, and generates coordinated whole-body actions via causal masked self-attention within a diffusion-based denoising process.
    
\subsection{Geometry-Adaptive Encoder}
\label{sec:encoder}
\textbf{Lightweight Fourier Spatial Encoding.}
To enrich visual semantic features with dense 3D spatial information without introducing a heavy encoder, we map PointMap observations into patch-level geometric tokens using a parameter-efficient Fourier MLP inspired by the observation that Fourier feature mappings enable networks to learn high-frequency functions from low-dimensional inputs~\citep{FourierFeatures}. At timestep $t$, the PointMap $P_t \in \mathbb{R}^{H \times W \times 3}$ is partitioned using the same patch layout as DINOv2 \citep{dinov2}, yielding $N$ spatially aligned patch regions. For each patch $i$, we select valid 3D points according to the depth validity mask and compute their mean coordinate as the patch-level geometric center $\bar{p}^i$. The geometric feature is then constructed as:
\begin{equation}
    F_g^i = \mathrm{MLP}\!\left(\left[\,\bar{p}^i;\; \psi(\bar{p}^i);\; v^i\,\right]\right),
\end{equation}
where $\psi(\cdot)$ denotes multi-band sinusoidal positional encoding, $v^i \in [0,1]$ is the ratio of valid depth points within patch $i$, and $[\,\cdot\,;\,\cdot\,]$ denotes concatenation. The raw coordinate $\bar{p}^i$ preserves low-frequency spatial location, the Fourier encoding $\psi(\bar{p}^i)$ resolves high-frequency geometric structure, and the validity ratio $v^i$ provides an explicit signal of depth reliability. Unlike ViT-based geometric encoders that require training from scratch with substantial parameters, our Fourier spatial encoder introduces only 1.4M parameters (see Table~\ref{tab:modality_fusion_ablation}) and is trained end-to-end with the policy, preserving real-time inference speed.

\textbf{Reliability-Aware Gated Fusion.}
Dense geometric features are valuable but inherently noisy: depth sensors produce invalid readings at reflective surfaces, object boundaries, and distant regions. Rather than injecting all geometric signals uniformly, we introduce a per-token scalar gate that modulates fusion strength based on feature compatibility and depth reliability:
\begin{equation}
    \gamma^i = \sigma\!\left(\mathbf{w}^\top \left[\,F_s^i;\; F_g^i;\; 
    v^i\,\right] + b\right),
\end{equation}
\begin{equation}
    F_v^i = F_s^i + \gamma^i \cdot F_g^i,
\end{equation}
where $F_s^i$ is the semantic feature from DINOv2~\citep{dinov2} with LoRA~\citep{LoRA} adaptation for patch $i$, $\sigma$ denotes the sigmoid function, and $\mathbf{w} \in \mathbb{R}^{2d+1}$, $b \in \mathbb{R}$ are learnable parameters. The gate $\gamma^i \in [0,1]$ is jointly conditioned on semantic context $F_s^i$, geometric content $F_g^i$, and depth validity $v^i$, enabling context-dependent fusion that goes beyond simple depth-quality filtering. As shown in Fig.~\ref{fig:overview_geohat}, the learned gate activates strongly on task-relevant regions where both geometry and semantics are informative, rather than uniformly following depth validity alone. This linear gating introduces only $2d{+}2$ parameters---negligible relative to the DINOv2 backbone---yet provides per-token adaptivity that a global scalar gate cannot achieve.

\subsection{Hybrid Whole-Body Action Decoder}
\label{sec:decoder}
Given the fused visual feature $F_v$, we propose a Hybrid Whole-Body Action Decoder to predict future whole-body actions. The action space over a prediction horizon $h$ is decomposed into two subspaces: the arm action $A^{\mathrm{arm}} \in \mathbb{R}^{h \times d_a}$, encompassing arm joints, gripper, torso, and head states, and the base action $A^{\mathrm{base}} \in \mathbb{R}^{h \times d_b}$, containing linear and angular velocities of the mobile base. During training, we sample a diffusion step $k$ and independently corrupt the ground-truth actions:
\begin{equation}
\begin{aligned}
    A_k^{\mathrm{arm}}
    &= \sqrt{\bar{\alpha}_k}\, A^{\mathrm{arm}}
    + \sqrt{1-\bar{\alpha}_k}\,\epsilon^{\mathrm{arm}}, \\
    A_k^{\mathrm{base}}
    &= \sqrt{\bar{\alpha}_k}\, A^{\mathrm{base}}
    + \sqrt{1-\bar{\alpha}_k}\,\epsilon^{\mathrm{base}},
\end{aligned}
\end{equation}
where $\bar{\alpha}_k$ is the cumulative noise schedule and $\epsilon^{\mathrm{arm}}, \epsilon^{\mathrm{base}} \sim \mathcal{N}(0,I)$.

\begin{wrapfigure}{r}{0.35\linewidth}
    \centering
    \includegraphics[width=1.0\linewidth]{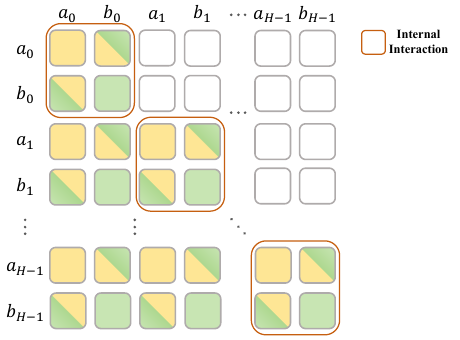}
    \vspace{-0.8em}
    \caption{Causal mask structure. Tokens within the same timestep attend bidirectionally, while cross-timestep attention is strictly causal.}
    \label{fig:causal_mask}
    \vspace{-0.1em}
\end{wrapfigure}

\textbf{Hybrid Action Rearrangement.} The noisy action sequences are projected into a shared latent space via separate projection layers. To encode temporal order and distinguish the two subspaces, we add a shared learnable temporal embedding and a subspace-specific embedding to each token:
\begin{equation}
\begin{aligned}
    z_i^{\mathrm{arm}}
    &= \phi_{\mathrm{arm}}(A_{k,i}^{\mathrm{arm}})
    + e_i^{\mathrm{temp}}
    + e^{\mathrm{arm}}, \\
    z_i^{\mathrm{base}}
    &= \phi_{\mathrm{base}}(A_{k,i}^{\mathrm{base}})
    + e_i^{\mathrm{temp}}
    + e^{\mathrm{base}},
\end{aligned}
\end{equation}
where $\phi_{\mathrm{arm}}$ and $\phi_{\mathrm{base}}$ are independent projection layers, $e_i^{\mathrm{temp}}$ is the temporal embedding for step $i$, and $e^{\mathrm{arm}}$, $e^{\mathrm{base}}$ are subspace-specific embeddings. The resulting tokens are interleaved temporally to form a hybrid action sequence:
\begin{equation}
    Z =
    \left[
    z_1^{\mathrm{arm}},\,
    z_1^{\mathrm{base}},\,
    z_2^{\mathrm{arm}},\,
    z_2^{\mathrm{base}},\,
    \dots,\,
    z_h^{\mathrm{arm}},\,
    z_h^{\mathrm{base}}
    \right].
\end{equation}
The interleaved ordering places arm and base tokens (denoted $a_i$ and $b_i$ in Fig.~\ref{fig:overview_geohat} and~\ref{fig:causal_mask} for brevity) of the same timestep adjacent in the sequence, enabling the causal mask (described below) to permit their mutual interaction with minimal positional distance.

\textbf{Whole-Body Action Decoder.} Each decoder layer is conditioned on both the diffusion step $k$ and the robot proprioceptive state $S_t$. The two embeddings are summed to form the decoder condition:
\begin{equation}
    c = \phi_t(k) + \phi_s(S_t),
\end{equation}
where $\phi_t$ and $\phi_s$ denote the timestep encoder and proprioception encoder, respectively. The condition $c$ modulates each decoder layer through Adaptive Layer Normalization ~\citep{AdaLN} (AdaLN), allowing feature transformations to adapt to the current denoising step and robot state. Each decoder layer comprises two mechanisms: query-level sparse attention for visual grounding and causal masked self-attention for temporal action modeling.

\textbf{Query-Level Sparse Attention.} To ground action generation in task-relevant visual context, hybrid action tokens $Z$ query the fused visual token $F_v$ pooled from all camera views. Rather than attending to all visual tokens uniformly, each action token independently selects its own top-$K$ most relevant visual tokens based on attention scores:
\begin{equation}
    \mathcal{S}_j = \frac{(z_j W_Q)(F_v W_K)^\top}{\sqrt{d_k}}, \quad
    \mathcal{I}_j = \operatorname{TopK}(\mathcal{S}_j,\, K),
\end{equation}
\begin{equation}
    \hat{z}_j = \operatorname{Softmax}\!\left(\mathcal{S}_j[\mathcal{I}_j]\right) \cdot (F_v[\mathcal{I}_j]\, W_V),
\end{equation}
where $W_Q, W_K, W_V$ are projection matrices, $\mathcal{I}_j$ is the index set of $K$ visual tokens selected for action token $z_j$, and $F_v[\mathcal{I}_j]$ denotes the corresponding subset. Since each action token selects independently, arm and base tokens naturally attend to different spatial regions according to their respective control demands---arm tokens tend to focus on contact-relevant object surfaces, while base tokens attend to broader spatial context for positioning (visualized in Fig.~\ref{fig:overview_geohat}). This query-level selection suppresses irrelevant background noise and improves both efficiency and robustness of visual-action alignment. We study the effect of $K$ in the ablation experiments.

\textbf{Causal Masked Self-Attention.} The action tokens are processed by causal masked self-attention to model temporal dependencies. For the interleaved sequence $Z$ of length $2h$, the attention mask $M \in \{0,1\}^{2h \times 2h}$ is defined as:
\begin{equation}
    M_{ij} = \mathbbm{1}\!\left[\left\lceil \tfrac{i}{2} \right\rceil \geq \left\lceil \tfrac{j}{2} \right\rceil\right],
\end{equation}
where $\lceil \cdot / 2 \rceil$ maps each token index to its timestep. As illustrated in Fig.~\ref{fig:causal_mask}, this mask permits bidirectional attention within each timestep (enabling arm-base coordination) while enforcing strict causality across timesteps (preventing future information leakage). After multiple decoder layers, the final tokens are projected back to arm and base action spaces via separate output heads.

\subsection{Training and Inference}

Following DP3~\citep{DP3}, we adopt a sample-prediction formulation where the denoising network $\mathcal{F}_{\theta}$ directly predicts clean actions rather than noise. Given clean actions $A^{\mathrm{arm}}$ and $A^{\mathrm{base}}$, both are independently perturbed according to the diffusion forward process, producing noisy actions $A_{k}^{\mathrm{arm}}$ and $A_{k}^{\mathrm{base}}$ at diffusion step $k$. Conditioned on the fused visual feature $F_v$, the proprioceptive state $S_t$, and the diffusion step $k$, the model predicts the corresponding clean action sequences. The training objective is:
\begin{equation}
    \mathcal{L}
    =
    \sum_{l\in\{\mathrm{arm},\mathrm{base}\}}
    \lambda_l \,
    \operatorname{MSE}
    \left(
    \hat{A}^{l},\,
    A^{l}
    \right), \quad
    \hat{A}^{l}
    =
    \mathcal{F}_{\theta}
    \left(
    A_{k}^{l},\, k,\, F_v,\, S_t
    \right),
\end{equation}
where we set $\lambda_{\mathrm{arm}} = \lambda_{\mathrm{base}} = 1$ in all experiments, more details are provided in Appendix~\ref{Appendix B}.

During inference, we adopt DDIM~\citep{DDIM} sampling with $T$ denoising steps for efficient generation. Starting from Gaussian noise $A_T^l \sim \mathcal{N}(0, I)$, the model iteratively predicts the clean sample $\hat{A}^l$ and updates the noisy actions according to the deterministic DDIM update rule until convergence. 
\section{Experiments}
\label{sec:experiments}
\subsection{Simulation Experiments} 
\noindent\textbf{Benchmark.}
We evaluate on ManiSkill-HAB (MSHAB)~\cite{maniskill}, built upon SAPIEN~\cite{sapien}, which provides diverse room layouts and interactive mobile manipulation tasks. Following the protocol of AC-DiT~\cite{AC-DiT}, 1,000 expert trajectories are collected per task using pretrained RL policies (PPO~\cite{PPO} for open/close, SAC~\cite{SAC} for pick/place). For all tasks except placement, only observations obtainable in real-world settings are used, excluding privileged information such as target poses and grasp success signals. Placement tasks additionally require privileged goal specification, as MSHAB defines their targets as invisible virtual positions that cannot be inferred from visual observations alone.

\noindent\textbf{Baselines.}
For fair comparison under a unified evaluation protocol, we adopt the results reported by AC-DiT~\cite{AC-DiT}, which re-implements and evaluates all baselines with standard data collection, preprocessing, and evaluation procedures on ManiSkill-HAB. For 2D RGB-based imitation learning, we include ACT~\cite{ACT} and DP~\cite{DP} as foundational baselines. DP3~\cite{DP3} extends the diffusion-based paradigm to 3D observation modeling. $\pi_0$~\cite{pi_0} and RDT-1B~\cite{rdt} provide scalable generative models for robot manipulation. AC-DiT integrates RGB observations, point clouds, and language instructions through a two-stage training paradigm.

\noindent\textbf{Evaluation Metrics.}
We report the task success rate as the primary evaluation metric. For each task, the policy is evaluated on 100 unseen episodes and the result is averaged over three trials (mean $\pm$ std). The success of each trajectory is determined according to the official task-specific success criteria defined in MSHAB~\cite{maniskill}. The maximum episode length is set to 200 steps for all task categories.
\begin{table}[t]
    \centering

    \renewcommand{\arraystretch}{1.3}
    \setlength{\tabcolsep}{6pt}

    \resizebox{1\linewidth}{!}{
    \begin{tabular}{lcccccccc}
        \toprule
        \multirow{4}{*}{\textbf{Method}} 
        & \multicolumn{7}{c}{\textbf{SetTable}} 
        & \multirow{4}{*}{\textbf{Avg}} \\
        \cmidrule(lr){2-8}

        & Pick 
        & Place
        & Open
        & Pick 
        & Place 
        & Open 
        & Close  
        &    \\
        & Apple 
        & Apple
        & Fridge
        & Bowl 
        & Bowl 
        & Drawer 
        & Drawer 

        &    \\
        
        \midrule
        ACT~\citep{ACT}        & $28.0\pm{2.2}$ & $8.7\pm{3.3}$ & $2.0\pm{2.2}$ & $28.0\pm{2.4}$ & $13.0\pm{0.8}$ & $0.0\pm{0.0}$ & $85.7\pm{1.2}$  & 23.6 \\
        DP~\citep{DP}         & $21.3\pm{3.3}$ & $28.0\pm{8.0}$ & $7.3\pm{5.8}$ & $20.7\pm{3.3}$ & $69.3\pm{3.3}$ & $0.0\pm{0.0}$ & $55.0\pm{5.7}$  & 28.8 \\
        DP3~\citep{DP3}        & $0.0\pm{0.0}$ & $31.0\pm{0.8}$ & $0.0\pm{0.0}$ & $20.0\pm{2.4}$ & $32.0\pm{0.8}$ & $0.0\pm{0.0}$ & $68.0\pm{0.0}$  & 21.6 \\
        $\pi_0$~\citep{pi_0} & $13.0\pm{1.6}$ & $23.3\pm{1.7}$ & $31.3\pm{2.6}$ & $15.7\pm{1.2}$ & $21.3\pm{2.6}$ & $60.0\pm{0.8}$ & $70.0\pm{2.2}$ & 33.5 \\
        RDT-1B~\citep{rdt}        & $12.0\pm{11.3}$ & $32.0\pm{5.7}$ & $82.7\pm{10.5}$ & $10.7\pm{6.8}$ & $18.7\pm{5.0}$ & $44.0\pm{8.6}$ & $\textbf{100.0}\pm{\textbf{0.0}}$ & 42.9 \\
        AC-DiT~\citep{AC-DiT} & $33.3\pm{1.9}$ & $33.3\pm{9.4}$ & $\textbf{90.7}\pm{\textbf{5.0}}$ & $36.0\pm{6.5}$ & $17.3\pm{6.8}$ & $81.3\pm{6.8}$ & $97.3\pm{1.9}$ & 55.6 \\
        \textbf{Ours}         & $\textbf{82.3}\pm{\textbf{2.5}}$ & $\textbf{60.0}\pm{\textbf{1.6}}$ & $83.7\pm{0.5}$ & $\textbf{69.3}\pm{\textbf{1.2}}$ & $\textbf{78.0}\pm{\textbf{0.0}}$ & $\textbf{86.0}\pm{\textbf{0.0}}$ & $95.3\pm{0.5}$ & $\textbf{79.2}\pm{\textbf{0.9}}$ \\
        \bottomrule
    \end{tabular}
    }
    \vspace{0.3em}
    \caption{Success rates on ManiSkill-HAB tasks. The results are reported on SetTable tasks. Additional experiments on TidyHouse and PrepareGroceries tasks are provided in Appendix~\ref{Appendix A.1}.}
    \label{tab:maniskill_hab_results}
    \vspace{-2.0em}
\end{table}

\noindent\textbf{Results.}
As shown in Table~\ref{tab:maniskill_hab_results}, GeoHAT achieves an average success rate of 79.2\% on SetTable tasks, outperforming all baselines by a significant margin. The improvement is particularly pronounced on pick and place tasks requiring precise base-arm coordination. In contrast, DP \cite{DP} and DP3~\cite{DP3} struggle across most tasks due to the difficulty of processing scene-level observations, while RDT-1B \cite{rdt} achieves competitive results on certain articulated tasks but falls short when tight base-arm coordination is required. Overall, GeoHAT maintains consistently strong performance across all task categories, demonstrating that geometry-adaptive gated fusion provides robust spatial reasoning while the hybrid action decoder enables tighter arm-base coordination.

\begin{table}[h]
    \centering
    \small
    \renewcommand{\arraystretch}{1.2}
    \setlength{\tabcolsep}{10pt}
    \begin{tabular}{lcccc}
        \toprule
        \textbf{Exp} & \textbf{Avg. SR} & $\boldsymbol{\Delta}\text{\textbf{SR}}$ & \textbf{Geo. Params (M)} & \textbf{Speed (ms)} \\
        \midrule
        \textbf{Ours}   & $\textbf{79.2}\pm{0.9}$ & --                      & \textbf{1.4} & \textbf{80} \\
        Addition        & $66.4\pm{0.8}$       & \textcolor{red}{-12.8}   & 6.7 & 85 \\
        Concatenation   & $72.1\pm{1.3}$       & \textcolor{red}{-7.1}   & 6.7  & 108 \\
        Joint Attention & $65.9\pm{1.9}$       & \textcolor{red}{-13.3}  & 13.1  & 123 \\
        \bottomrule
    \end{tabular}
    \vspace{0.3em}
    \caption{Ablation on geometry fusion strategies. Geo.\ Params: parameters of the geometric encoder. Speed: per-step inference time (ms). Our gated fusion performs the best with the fewest parameters.}
    \label{tab:modality_fusion_ablation}
    \vspace{-2.5em}
\end{table}
\subsection{Ablation Study}

\textbf{Geometry-Adaptive Fusion.} Table~\ref{tab:modality_fusion_ablation} compares different strategies for integrating PointMap features with RGB representations. Following PointMapPolicy~\cite{PointMapPolicy}, we encode PointMap observations with a ViT~\cite{ViT} trained from scratch and fuse them via addition or concatenation, achieving 66.4\% and 72.1\% success rates, respectively. Joint attention fusion further introduces a cross-modal attention module (adding 6.4M parameters beyond the ViT encoder), yet degrades performance to 65.9\%, suggesting that overly coupled fusion disrupts the pretrained RGB features. Our gated fusion achieves the best result of 79.2\% while requiring only 1.4M geometric parameters---4.8$\times$ fewer than ViT-based alternatives---and the fastest inference speed. The efficiency advantage is twofold: the Fourier encoder replaces the ViT forward pass with a lightweight MLP, and gated fusion preserves the original token count in subsequent cross-attention, unlike concatenation which doubles the key-value sequence length. These results show that our reliability-aware fusion achieves stronger geometric grounding with fewer parameters and lower latency than fixed fusion alternatives.

\begin{wrapfigure}{r}{0.40\linewidth}
    \centering
    \vspace{-1.0em}
    \resizebox{\linewidth}{!}{
        \renewcommand{\arraystretch}{1.2}
\setlength{\tabcolsep}{14pt}

\begin{tabular}{lcc}
    \toprule
    \textbf{Exp} & \textbf{Avg} & $\boldsymbol{\Delta}$ \\
    \midrule
    \textbf{Ours}   & $\textbf{79.2}\pm{\textbf{0.9}}$ & -- \\
    Single Stream   & $70.8\pm{1.2}$ &  \textcolor{red}{-8.4} \\
    Ordered Stream & $70.3\pm{1.5}$ & \textcolor{red}{-8.9} \\
    \bottomrule
\end{tabular}
    }
    \vspace{-0.5em}
    \captionof{table}{Ablation study on different action representations.}
    \label{tab:action_progress}
    \vspace{-1.0em}
\end{wrapfigure}
\textbf{Hybrid Action Representation.}
Table~\ref{tab:action_progress} reports the ablation on whole-body action design. A unified action sequence (Single Stream, following DSPv2~\cite{DSPv2}) underperforms because joint training lets high-DOF arm features dominate the low-DOF base signal. Conditioning arm on base predictions (Ordered Stream, following WB-VIMA~\cite{WB-VIMA}) also degrades performance, as slight base errors propagate to and amplify in the arm. Our interleaved decomposition achieves the best results, effectively balancing both subsystems' dynamics while enabling tighter whole-body coordination.

\begin{wrapfigure}{r}{0.35\linewidth}
    \centering
    \small
    \renewcommand{\arraystretch}{1.2}
\setlength{\tabcolsep}{10pt}
\begin{tabular}{lcc}
    \toprule
    \textbf{\(K\)} & \textbf{Avg} & $\boldsymbol{\Delta}$ \\
    \midrule
    \textbf{64}   & $\textbf{79.2}\pm{\textbf{0.9}}$ & -- \\
    32   & $75.2\pm{1.4}$          &  \textcolor{red}{-4.0} \\
    Full & $73.4\pm{1.1}$          &  \textcolor{red}{-5.8} \\
    \bottomrule
\end{tabular}
    \vspace{-0.5em}
    \captionof{table}{Ablation study on Top-$K$ Selection.}
    \label{tab:topk_selection}
    \vspace{-1.0em}
\end{wrapfigure}
\textbf{Top-$K$ Selection.}
We study the effect of $K$ in query-level sparse attention. Full attention ($K{=}\text{all}$) yields 73.4\%, a 5.8~pp drop from our default $K{=}64$, confirming that attending to all visual tokens introduces irrelevant background noise. Reducing to $K{=}32$ also degrades performance to 75.2\%, as overly aggressive filtering removes useful spatial context. The optimal $K{=}64$ provides a balanced trade-off, retaining sufficient visual tokens to cover both local contact regions for arm actions and broader scene context for base positioning. Additional ablation studies on the generality of our geometry encoder and robustness to depth perturbations are provided in Appendix~\ref{Appendix A.2} and~\ref{Appendix A.3}.

\subsection{Real-World Experiments}
\begin{wrapfigure}{r}{0.35\linewidth}
    \vspace{-1.0em}
    \centering
    \includegraphics[width=1.0\linewidth]{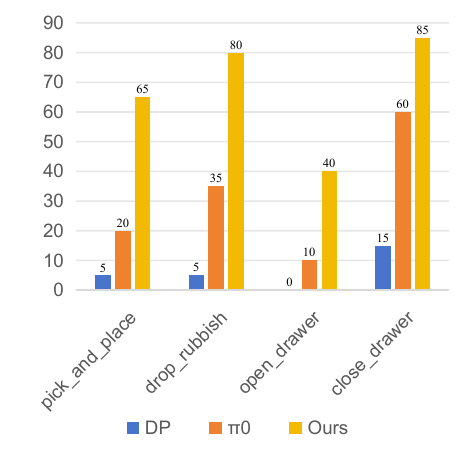}
    \vspace{-2.5em}
    \caption{Success rate of GeoHAT and baselines in real-world tasks.}
    \label{fig:real_world_result}
    \vspace{-1em}
\end{wrapfigure}
To further validate real-world applicability, we conduct experiments on four mobile manipulation tasks requiring coordinated base-arm control: \textit{pick\_and\_place}, \textit{drop\_rubbish}, \textit{open\_drawer}, and \textit{close\_drawer}. Our platform consists of an AIRBOT Play robot arm, a Woosh mobile base, and two Intel RealSense D455 cameras placed on the head and wrist. Additional hardware details are provided in Appendix~\ref{Appendix C}. For each task, we collect 50 teleoperated demonstrations. We compare quantitatively against DP~\cite{DP} and $\pi_0$~\cite{pi_0}, evaluating each method over 20 trials per task under identical conditions. Notably, both baselines operate on RGB only, as incorporating depth in mobile settings is non-trivial: consumer-grade sensors exhibit degraded readings under shifting viewpoints and reflective surfaces, and 3D methods~\citep{DSPv2,DP3} require extrinsic calibration that complicates deployment. GeoHAT addresses these issues through calibration-free PointMap processing in each camera's local frame and reliability-aware gating that suppresses unreliable regions, enabling effective depth utilization where prior methods opt to discard it. As shown in Fig.~\ref{fig:real_world_result}, GeoHAT consistently outperforms all baselines across every task. Task definitions and execution details are provided in Appendix~\ref{Appendix D}.

\section{Conclusion}
\label{sec:conclusion}

We presented GeoHAT, a diffusion-based framework for whole-body mobile manipulation that combines reliable geometric perception with coordinated action generation. A lightweight Fourier encoder and reliability-aware gating enrich pretrained visual features with dense 3D information, while a hybrid action decoder uses subspace-specific sparse attention and causal self-attention to coordinate arm and base actions. Experiments on ManiSkill-HAB and real-world tasks demonstrate improvements over representative baselines, with ablations supporting the contribution of each component. These results highlight the value of reliability-aware geometric fusion and action-specific visual conditioning for mobile manipulation.

\section{Limitations}
\label{sec:limitations}

GeoHAT is bounded by demonstration coverage; tasks with rare coordination patterns may lead to failures. Our simulation evaluation focuses on SetTable tasks, which cover pick, place, open, and close primitives spanning the core base-arm coordination modes, but broader categories remain to be validated. Although the reliability-aware gate suppresses noisy depth patches, systematic failures on fully reflective or transparent surfaces persist. Task-specific training without vision-language pretraining constrains generalization to unseen objects or novel instructions. Extending the framework with reinforcement learning, VLA representations, and hierarchical planning are promising future directions.

\clearpage
\acknowledgments{
We sincerely thank the anonymous reviewers for their valuable comments and suggestions, which substantially improved the completeness of this paper and encouraged us to examine the underlying methodological and experimental details more thoroughly. From the initial conception of the idea to the completion of extensive simulation and real-world experiments, this work was made possible by the close collaboration, mutual support, and
sustained dedication of the three co-first authors. We also express our sincere gratitude to our two faculty mentors for their insightful guidance and valuable advice throughout
the project.
}


\bibliography{main}  
\clearpage
\newpage
\appendix
\onecolumn

\begin{center}
    {\LARGE \textbf{Appendix}}
\end{center}
\vspace{0.5em}

\section{Additional Experiments}
\label{Appendix A}

\subsection{Complex Experiments on MSHAB}
\label{Appendix A.1}
 \begin{table}[h]
      \centering
      \small
      \renewcommand{\arraystretch}{1.2}
      \setlength{\tabcolsep}{8pt}

      \begin{tabular}{lcccccc}
          \toprule
          \multirow{2}{*}{\textbf{Method}}
          & \multicolumn{3}{c}{\textbf{TidyHouse}}
          & \multicolumn{3}{c}{\textbf{PrepareGroceries}} \\
          \cmidrule(lr){2-4} \cmidrule(lr){5-7}
          & Pick\_All
          & Place\_All
          & Avg
          & Pick\_All
          & Place\_All
          & Avg \\
          \midrule
          ACT & $0.0\pm{0.0}$ & $58.0\pm{0.0}$ & 29.0 & $0.0\pm{0.0}$ & $34.0\pm{0.0}$ & 17.0 \\
          DP   & $0.7\pm{0.5}$ & $32.7\pm{0.5}$ & 16.7 & $0.0\pm{0.0}$ & $29.7\pm{1.3}$ & 14.9 \\
          DP3 & $0.0\pm{0.0}$ & $61.0\pm{0.0}$ & 30.5 & $0.0\pm{0.0}$ & $31.3\pm{1.2}$ & 15.7 \\
          \textbf{Ours}   & $\textbf{30.3}\pm{\textbf{0.5}}$ & $\textbf{73.3}\pm{\textbf{1.7}}$ & $\textbf{51.8}\pm{\textbf{1.1}}$ & $\textbf{19.7}\pm{\textbf{4.6}}$ & $\textbf{60.7}\pm{\textbf{2.5}}$ & $\textbf{40.2}\pm{\textbf{3.6}}$ \\
          \bottomrule
      \end{tabular}

      \vspace{0.5em}
      \caption{Success rates on ManiSkill-HAB TidyHouse and PrepareGroceries tasks.}
      \label{tab:tidyhouse and preparegroceries results}
      \vspace{-1.5em}
  \end{table}

To further assess our approach in more complex and diverse environments, we evaluate it on the TidyHouse and PrepareGroceries tasks from ManiSkill-HAB. Each task is evaluated under both the Pick All and Place All settings, which involve 9 target objects distributed across different workspaces; a single policy is trained to handle all 9 objects per setting. We compare against three baselines: ACT and DP with 2D RGB input, and DP3 with 3D point-cloud input. For fair evaluation, we test on 100 episodes unseen during training, conduct three independent runs per task, and report the mean and standard deviation. As shown in Table~\ref{tab:tidyhouse and preparegroceries results}, our method surpasses all baselines on every task, achieving consistently high success rates.
 
We follow the same training protocol as in the SetTable experiments. For Pick All tasks, we use only non-privileged states: robot joint angles, joint velocities, and TCP pose. For Place All tasks, since the placement target is a visually invisible virtual point, we additionally provide the target object pose and the base target pose, following MSHAB conventions. We collect 1{,}000 trajectories per target object, pool the data across all object categories within each scene, and train for 100k timesteps with a batch size of 64. During evaluation, the target object in each episode is selected at random, which substantially increases task difficulty. To illustrate how our model performs in MSHAB, we present the execution process for each task in the SetTable, TidyHouse, and PrepareGroceries environments in Figures~\ref{fig:set_table}, ~\ref{fig:tidy_house}, and ~\ref{fig:prepare_groceries}, respectively. As shown, all tasks in MSHAB require precise coordination between the mobile base and the arm.

\subsection{Ablation Studies on the Generality of Our Geometry Encoder}
\label{Appendix A.2}
\begin{center}
    \centering
    \small
        \renewcommand{\arraystretch}{1.2}
\setlength{\tabcolsep}{14pt}

\begin{tabular}{lc}
    \toprule
    \textbf{Exp} & \textbf{Avg} \\
    \midrule
    \textbf{Ours w/Pointmap} & $\textbf{79.2}\pm{\textbf{0.9}}$ \\
    \midrule
    Ours w/depth             & $72.4\pm{1.1}$ \\
    \quad - Addition       & $51.0\pm{0.7}$ \\
    \quad - Concatenation  & $64.6\pm{1.2}$ \\
    \quad - Joint Attention & $67.8\pm{1.7}$ \\
    \midrule
    Ours w/DAv2              & $72.8\pm{0.7}$ \\
    \bottomrule
\end{tabular}

        \vspace{0.2em}
        \captionof{table}{Ablation study on geometric input modality.}
        \label{tab:depth_ablation}
\end{center}

To validate that our lightweight spatial encoder generalizes to the depth modality, we conduct experiments using the same three fusion strategies employed in PointMapPolicy and compare them against our approach. We additionally include a baseline that encodes depth with Depth~Anything~V2\citep{yang2024depth} (DAv2) using a ViT-Base backbone---matching the architecture scale of our DINOv2 encoder. This comparison demonstrates that our method achieves comparable performance without relying on a heavy pretrained depth encoder.
 
To illustrate generalization across different 3D modalities, we perform ablation studies on the ManiSkill-HAB SetTable tasks. As shown in Table~\ref{tab:depth_ablation}, the Addition fusion strategy yields only 51.0\% on average, indicating that directly adding ViT-encoded depth features to pretrained RGB tokens severely disrupts the learned representations. Switching to Concatenation improves performance to 64.6\%, as depth tokens remain separate from RGB tokens, allowing query-level sparse attention to selectively down-weight uninformative geometric features. Joint Attention further improves performance to 67.8\% by introducing a dedicated cross-modal interaction module. However, because this depth encoder is trained from scratch, its features still lack the maturity to fully complement the pretrained visual priors, leaving a 4.6~pp gap relative to our method.
 
When PointMap is replaced with monocular depth encoded by a pretrained DAv2 backbone, the average success rate reaches 72.8\%, within 0.4~pp of our depth variant (72.4\%). This indicates that the gated fusion mechanism is agnostic to the choice of geometric encoder: a lightweight Fourier MLP with 1.4M parameters matches this performance without the 86.6M-parameter DAv2 backbone.
 
\subsection{Ablation Studies on Depth Perturbations}
\label{Appendix A.3}

\begin{table}[h]
    \centering
    \small
    \renewcommand{\arraystretch}{1.2}
    \setlength{\tabcolsep}{10pt}

    \begin{tabular}{lccccccc}
        \toprule
        \multirow{2}{*}{\textbf{Method}} &
        \multirow{2}{*}{\textbf{Origin}} &
        \multicolumn{3}{c}{\textbf{Gaussian Noise} ($\boldsymbol{\Delta}$) $\downarrow$} &
        \multicolumn{3}{c}{\textbf{Depth Dropout} ($\boldsymbol{\Delta}$) $\downarrow$} \\
        \cmidrule(lr){3-5} \cmidrule(lr){6-8}
        & & $100\times$ & $200\times$ & $500\times$ & 10\% & 20\% & 50\% \\
        \midrule
        \textbf{Ours} & 79.2 & 0.95 & 1.74 & \textcolor{green}{1.70} & \textcolor{green}{1.10} & \textcolor{green}{1.05} & \textcolor{green}{1.32} \\
        Ours w/DAv2   & 72.8 & \textcolor{green}{0.33} & \textcolor{green}{1.41} & 3.19 & 1.52 & 6.19 & 8.60 \\
        \bottomrule
    \end{tabular}
    \vspace{1.0em}
    \caption{Ablation study on depth perturbations during evaluation. The first row corresponds to our method. The second row shows results obtained by replacing the PointMap with a depth map encoded via Depth Anything v2, with features fused through addition.}
    \vspace{-2.0em}
    \label{tab:depth_perturbations_during_evaluation}
\end{table}
Since real-world environments commonly suffer from incomplete and inaccurate depth, we validate the robustness of our method against depth perturbations. As shown in Table~\ref{tab:depth_perturbations_during_evaluation}, we perturb depth observations in the SetTable environment using two strategies---additive Gaussian noise and random depth dropout, and evaluate model performance at test time. For each set of experiments, we evaluate on 100 unseen episodes during training and report the average over three runs.

\begin{figure}[h]
    \centering
    \includegraphics[width=0.95\linewidth]
    {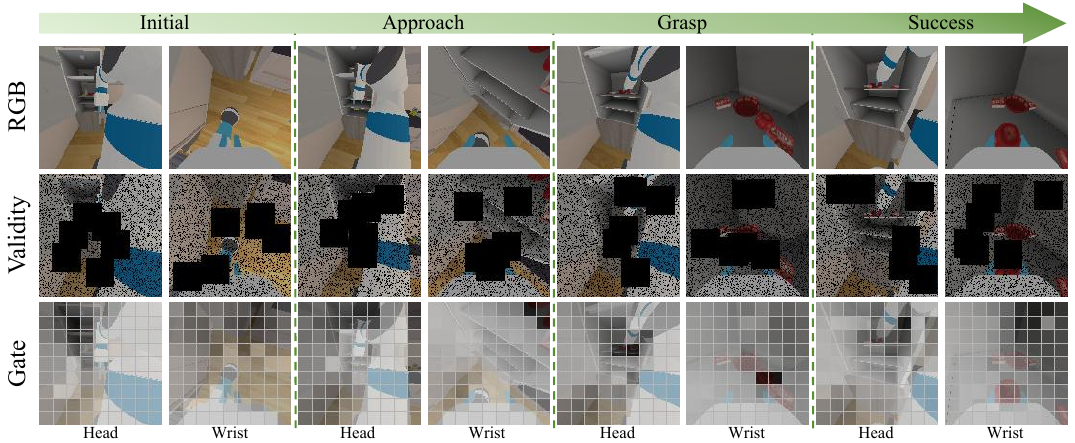}
    \caption{Gate vis. with task progress and depth corruption. The black square is used to simulate large-scale missing depth, while the remaining black dots simulate inaccurate depth.}
    \vspace{-1.0em}
    \label{fig:gate_vis}
\end{figure}

For Gaussian noise experiments, we corrupt raw depth values (in mm) with random noise at magnitudes of 100, 200, and 500. Under all noise levels, our method exhibits minimal performance degradation ($\Delta \leq 1.74$), demonstrating strong robustness to depth inaccuracy. For depth dropout experiments, we randomly mask 10\%, 20\%, and 50\% of depth values over randomly selected regions. Even at 50\% dropout, our method degrades by only 1.32 points, whereas the DAv2 baseline drops by 8.60 points. This gap confirms that the per-token reliability-aware gate---which receives the patch-level validity ratio $v^i$ as an explicit input---actively suppresses tokens from corrupted regions, providing robust geometric fusion for real-world deployment where depth quality is unreliable. In Fig.~\ref{fig:gate_vis}, we visualize $\gamma^i$ across four task stages and noisy depth quality.

\subsection{Visualizations of Query-level Sparse Attention}
To further validate the effectiveness of Query-level Sparse Attention, we provide supplementary visualizations corresponding to Figure~2, as shown in Figure~\ref{fig:attention_map}. We visualize how the attention regions attended by the arm and base evolve throughout task progression in the Pick Bowl task under the SetTable configuration. In the early phase, both the arm and base attend to task-relevant regions as well as their respective body locations. As the arm approaches the bowl, its attention concentrates on the bowl itself, while the base attends to the surrounding environment to assist manipulation. After a successful grasp, the arm attends solely to the grasped object, while the base continues monitoring the surrounding environment to maintain a stationary pose.

\begin{center}
    \vspace{0.5em}
    \begin{minipage}{\linewidth}
    \centering
    \includegraphics[width=1.0\linewidth]{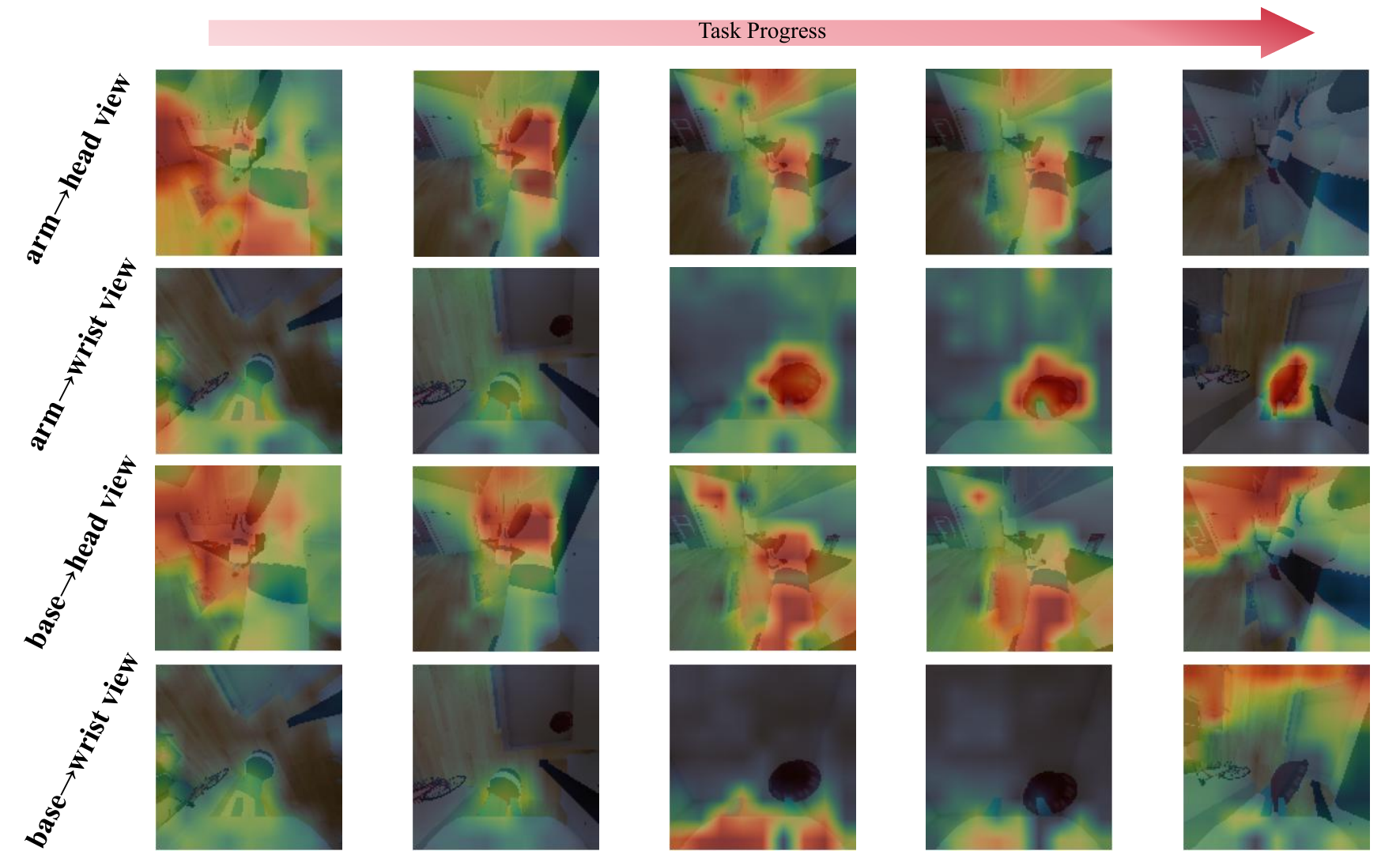}
    \captionof{figure}{Query-level Sparse Attention Results of Pick Bowl Task.}
    \label{fig:attention_map}
    \vspace{0.5em}
    \end{minipage}
\end{center}

\section{Implementation Details}
\label{Appendix B}

\noindent\textbf{Network Architecture.} We use DINOv2-Base (ViT-B/14, hidden size 768) as the visual backbone, fine-tuned with LoRA (rank 16, dropout 0.1). Input RGB images are resized and cropped to $126 \times 126$, yielding $9 \times 9 = 81$ patch tokens per view. The backbone output is projected from 768 to the model hidden dimension $d = 512$ via a linear layer. Our Fourier spatial encoder uses frequency bands $B = 4$. The Hybrid Whole-Body Action Decoder consists of 24 DiT blocks, each containing: (1) an AdaLN-modulated feed-forward layer, (2) query-level sparse attention (Top-$K{=}64$) over the fused visual tokens, (3) causal masked self-attention over the interleaved action sequence, and (4) a second AdaLN-modulated feed-forward layer. All attention modules use 8 heads. The MLP ratio is 4.0 (hidden dimension 2048). The arm action space has $d_a = 11$ dimensions (7 arm joints, 1 gripper, 1 torso lift, 2 head joints) and the base action space has $d_b = 2$ dimensions (linear and angular velocity). The prediction horizon is $h = 16$ with an action execution horizon of 8 steps. The proprioceptive state (31 dimensions, normalized to $[-1, 1]$) is encoded by a two-layer MLP with hidden dimension 2048 and summed with the sinusoidal diffusion timestep embedding to form the AdaLN condition.

\vspace{-0.5em}
\begin{center}
    \small
    \renewcommand{\arraystretch}{1.2}
    \setlength{\tabcolsep}{10pt}
    \begin{tabular}{lc}
        \toprule
        \textbf{Hyperparameter} & \textbf{Value} \\
        \midrule
        Batch Size & 64 \\
        Optimizer & AdamW \\
        Weight Decay & $1\times10^{-6}$ \\
        Learning Rate & $1\times10^{-4}$ \\
        Learning Rate Scheduler & Cosine Scheduler \\
        Learning Rate Warmup Steps & 500 \\
        EMA Power & 0.75 \\
        Action Chunk & 16 \\
        Top-$K$ & 64 \\
        Decoder Layers & 24 \\
        \bottomrule
    \end{tabular}
    \captionof{table}{Training hyperparameters.}
    \label{tab:training_params}
    \vspace{-0.5em}

\end{center}

\noindent\textbf{Diffusion Configuration.}
We use a squared cosine beta schedule with 100 training diffusion steps and sample-prediction (predicting $\hat{x}_0$ directly). During inference, DDIM sampling with 10 denoising steps is used for efficient generation.

\noindent\textbf{Training Details.}
All models are trained on a single NVIDIA A100 GPU for 30k iterations per task. We use AdamW optimizer with a cosine learning rate schedule (warmup 500 steps) and apply exponential moving average (EMA) with power 0.75 for stable convergence. During inference, we employ DDIM as the noise scheduler with 10 denoising steps for efficient action generation. Table~\ref{tab:training_params} summarizes the full set of training hyperparameters.

\noindent\textbf{Data Collection.}
For simulation experiments on ManiSkill-HAB, we collect 1,000 expert trajectories per task using pretrained RL policies: PPO for open/close tasks and SAC for pick/place tasks. Each trajectory records dual-view RGB images ($128 \times 128$) and depth maps from head and wrist cameras, along with proprioceptive states and whole-body actions. Depth maps are converted to PointMaps in the respective camera coordinate frames using known camera intrinsics, with a valid depth range of $(0, 3]$\,m. PointMap coordinates are scaled by a factor of 0.2 to normalize the spatial range. Trajectories are truncated at the first success step to remove redundant post-success frames. For real-world experiments, we collect 50 teleoperated demonstrations per task at the same control frequency.

\section{Robot Hardware Details}
\label{Appendix C}
\begin{center}
    \centering
    \begin{minipage}[t]{0.46\textwidth}
        \centering
        \begin{tabular}{cc}
            \toprule
            \textbf{Joint Name} & \textbf{Range} \\
            \midrule
            J1 & $-180^\circ \sim 120^\circ$ \\
            J2 & $-170^\circ \sim 10^\circ$ \\
            J3 & $-5^\circ \sim 180^\circ$ \\
            J4 & $-172.5^\circ \sim 172.5^\circ$ \\
            J5 & $-105^\circ \sim 105^\circ$ \\
            J6 & $-172^\circ \sim 172^\circ$ \\
            \bottomrule
        \end{tabular}
        \vspace{0.5em}
        \captionof{table}{AIRBOT Play Arm Position Ranges}
        \label{tab:arm_range}
    \end{minipage}
    \hfill
    \begin{minipage}[t]{0.46\textwidth}
        \centering
        \begin{tabular}{cc}
            \toprule
            \textbf{Parameter} & \textbf{Value} \\
            \midrule
            \multicolumn{2}{c}{\textbf{Intel RealSense D455}} \\
            \midrule
            Stream & RGB-D aligned \\
            Resolution & $640 \times 480$ \\
            Frequency & 30 FPS \\
            \bottomrule
        \end{tabular}
        \vspace{0.5em}
        \captionof{table}{Configurations for cameras}
        \label{tab:camera_config}
    \end{minipage}
\end{center}

\begin{center}
    \begin{minipage}{\linewidth}
    \centering
    \includegraphics[width=1.0\linewidth]{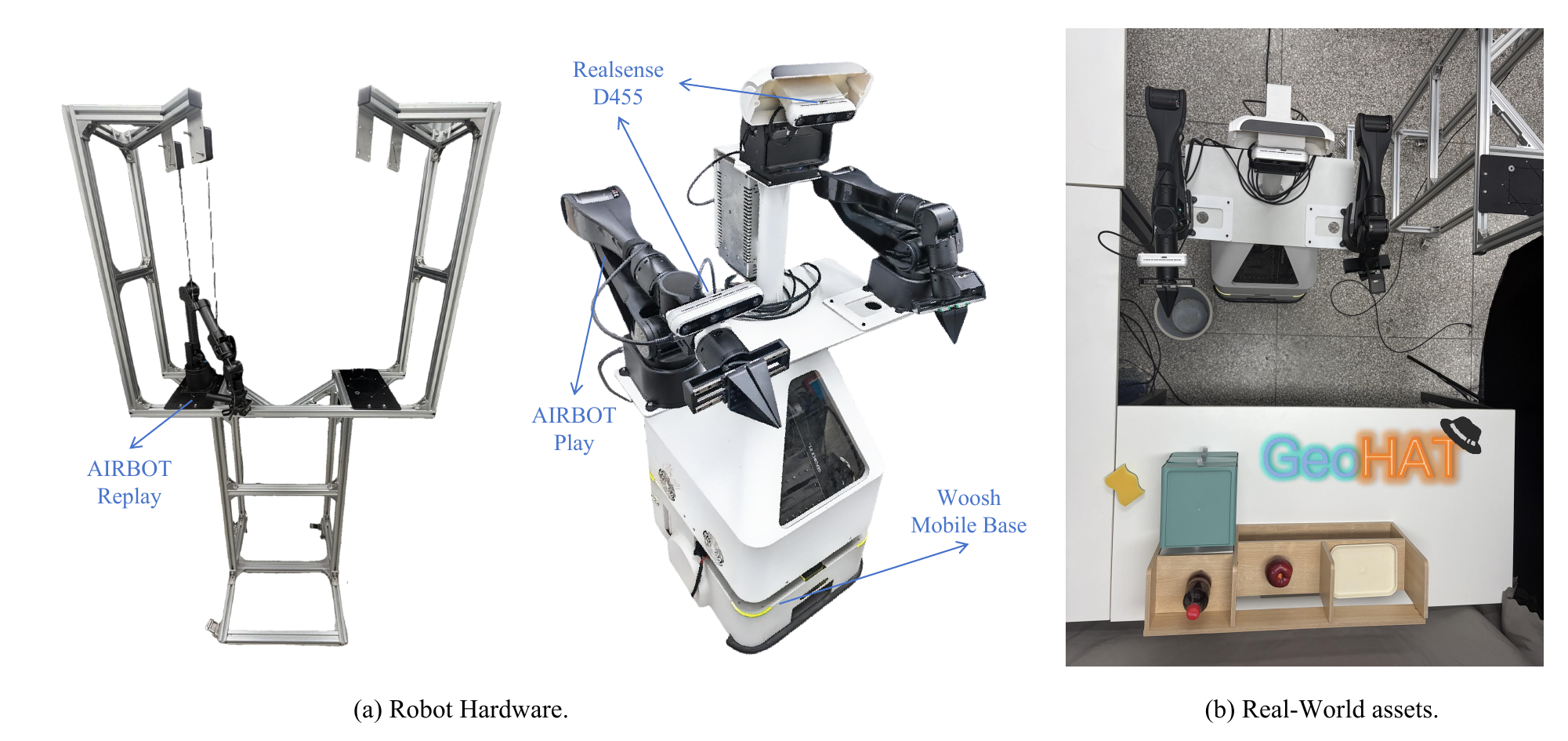}
    \captionof{figure}{Real-world robot hardware configuration. The mobile manipulation platform is equipped with an AIRBOT Play robotic arm, a Woosh mobile base, and RGB-D cameras mounted on the head and wrist for visual and geometric observations.}
    \label{fig:real_world_config}
    \end{minipage}
\end{center}

\textbf{Arms \& mobile base.} As illustrated in Fig.~\ref{fig:real_world_config} (a), our mobile manipulation platform uses an AIRBOT Replay arm as the master arm for data collection, while an AIRBOT Play arm serves as the slave arm during teleoperation and performs manipulation during the execution of real-world tasks. The AIRBOT Play arm is equipped with a parallel gripper with an opening range of 0--60 mm and operates in joint position control mode; the motion range of each joint is detailed in Table~\ref{tab:arm_range}. The mobile base is a Woosh wheeled platform that supports linear and angular velocity control. During both data collection and policy execution, we limit the maximum linear velocity to 0.2 m/s and the maximum angular velocity to 0.3 rad/s for stable and safe mobile manipulation.

\textbf{Cameras.} As illustrated in Fig.~\ref{fig:real_world_config} (a), our platform uses Intel RealSense D455 RGB-D cameras mounted on the head and wrist to provide visual and geometric observations. Each camera streams RGB-D aligned images at a resolution of $640 \times 480$ and 30 FPS, as summarized in Table~\ref{tab:camera_config}. GeoHAT processes observations from each camera in its own camera coordinate frame and does not require calibrated camera extrinsics. This calibration-free design reduces deployment overhead and makes the policy easy to transfer, enabling rapid adaptation to robots with different camera placements or hardware configurations.

\section{Real-World Experiment Details }
\label{Appendix D}

\begin{figure}[h]
    \centering
    \includegraphics[width=1.0\linewidth]{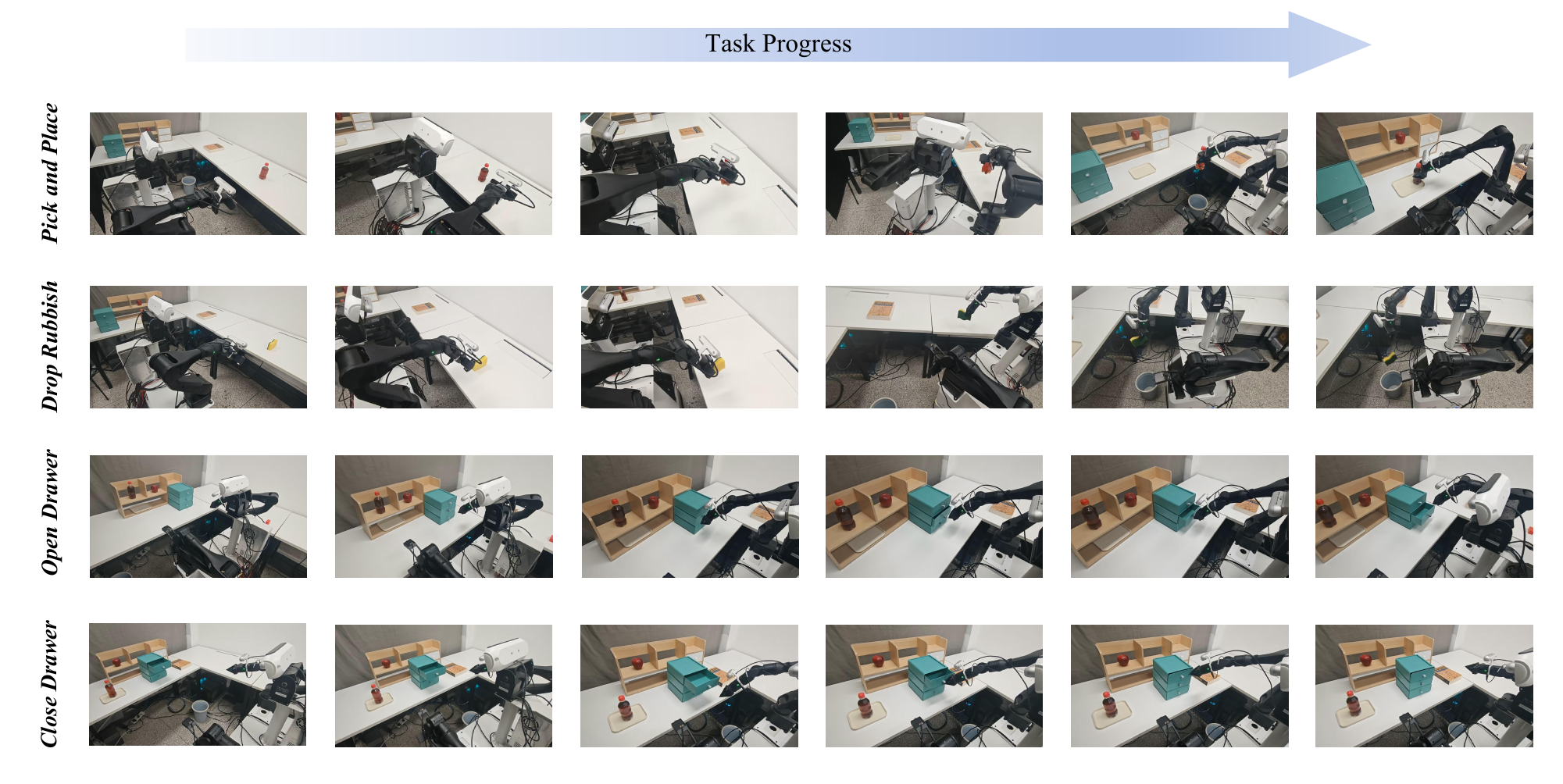}
    \caption{Execution progress of real-world mobile manipulation tasks. Each row shows representative temporal snapshots for one task, illustrating how the robot coordinates base motion and arm manipulation to complete pick-and-place, drop rubbish, open drawer, and close drawer.}
    \label{fig:real_world_tasks}
\end{figure}

During real-world deployment, GeoHAT uses aligned RGB-D observations from the wrist and head cameras, and converts the aligned depth maps into PointMaps for geometry-aware policy inference, as shown in Fig.~\ref{fig:real_world_obs}. Because the platform does not have torso and head joints, our policy is executed at 10 Hz and outputs a 9D action vector consisting of six arm joint targets, one gripper target, and two base velocity commands. In contrast, all baselines use only RGB images as their visual input. Fig.~\ref{fig:real_world_tasks} provides a detailed visualization of the execution progress for each real-world task. We evaluate GeoHAT and all baselines on four challenging mobile manipulation tasks, described as follows:

\begin{center}
    \begin{minipage}{\linewidth}
    \centering
    \vspace{-0.5em}
    \includegraphics[width=1.0\linewidth]{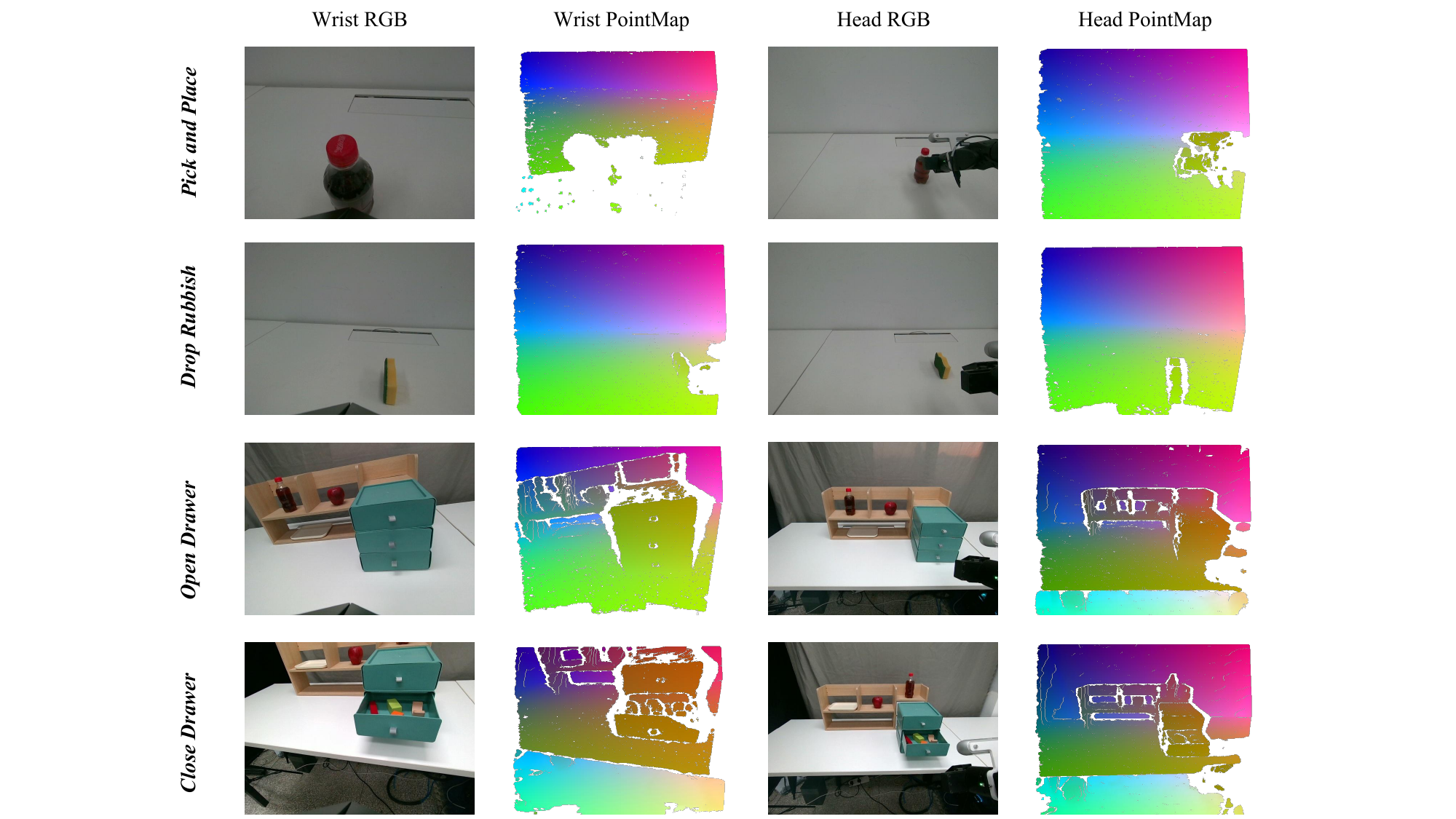}
    \captionof{figure}{Real-world observations used by GeoHAT. For each task, we show synchronized wrist-view and head-view RGB images together with their corresponding PointMaps converted from aligned depth observations. The PointMap representation provides dense geometric cues in each camera coordinate frame without requiring camera extrinsic calibration.}
    \label{fig:real_world_obs}
    \end{minipage}
\end{center}

\textbf{Pick-and-place.} The robot starts in front of a tabletop with a cola can placed on the table. The target placement region is a beige plastic tray located on the table to the left of the robot's initial position. A trial is considered successful if the robot picks up the cola can, places it into the tray, and the can remains stable without falling after placement.

\textbf{Drop rubbish.} The robot starts in front of a tabletop with a sponge used as the target rubbish. The rubbish bin is placed on the floor to the left of the robot's initial position. Success requires the robot to pick up the sponge from the table and accurately drop it into the rubbish bin.

\textbf{Open drawer.} This is the most challenging task among the four real-world evaluations. The robot starts in front of a table with a drawer that is initially closed. A trial is considered successful if the robot moves to the drawer, accurately grasps the small handle, and fully opens it. Since the handle is small and the drawer starts from a closed state, this task requires precise visual perception, accurate base positioning, and fine-grained manipulation, making it the most demanding task in our real-world benchmark.

\textbf{Close drawer.} The robot starts facing a tabletop drawer that is initially half open. The task is successful if the robot moves to the drawer and fully closes it.

\section{Failure Case Analysis}

Through extensive evaluation across simulation and real-world settings, we observe three recurring failure patterns. (1)~\textit{Sequential coordination errors.} In multi-phase tasks such as pick-and-place, suboptimal grasping poses in the first phase propagate to subsequent placement, causing minor positional deviations that ultimately lead to task failure. Improving demonstration quality---particularly ensuring diverse and precise grasp configurations---can alleviate this issue. (2)~\textit{Locomotion instability.} Since the mobile base operates under velocity control, where actual displacement results from integrating velocity commands over time, abrupt velocity jumps at any timestep can destabilize the controller, causing progressive drift that accumulates into unrecoverable errors. Standardizing the data collection protocol and maintaining numerically stable velocity profiles during teleoperation can mitigate this failure mode. (3)~\textit{Target disambiguation.} When lighting conditions or background appearances deviate from the training distribution, or in tasks such as TidyHouse that involve multiple semantically similar objects, the policy may select incorrect target objects due to the absence of explicit language guidance. Incorporating natural language instructions as an additional conditioning signal is a promising direction to resolve such ambiguities.

\clearpage

\begin{center}
    \begin{minipage}{\linewidth}
    \centering
    \includegraphics[width=1.0\linewidth]{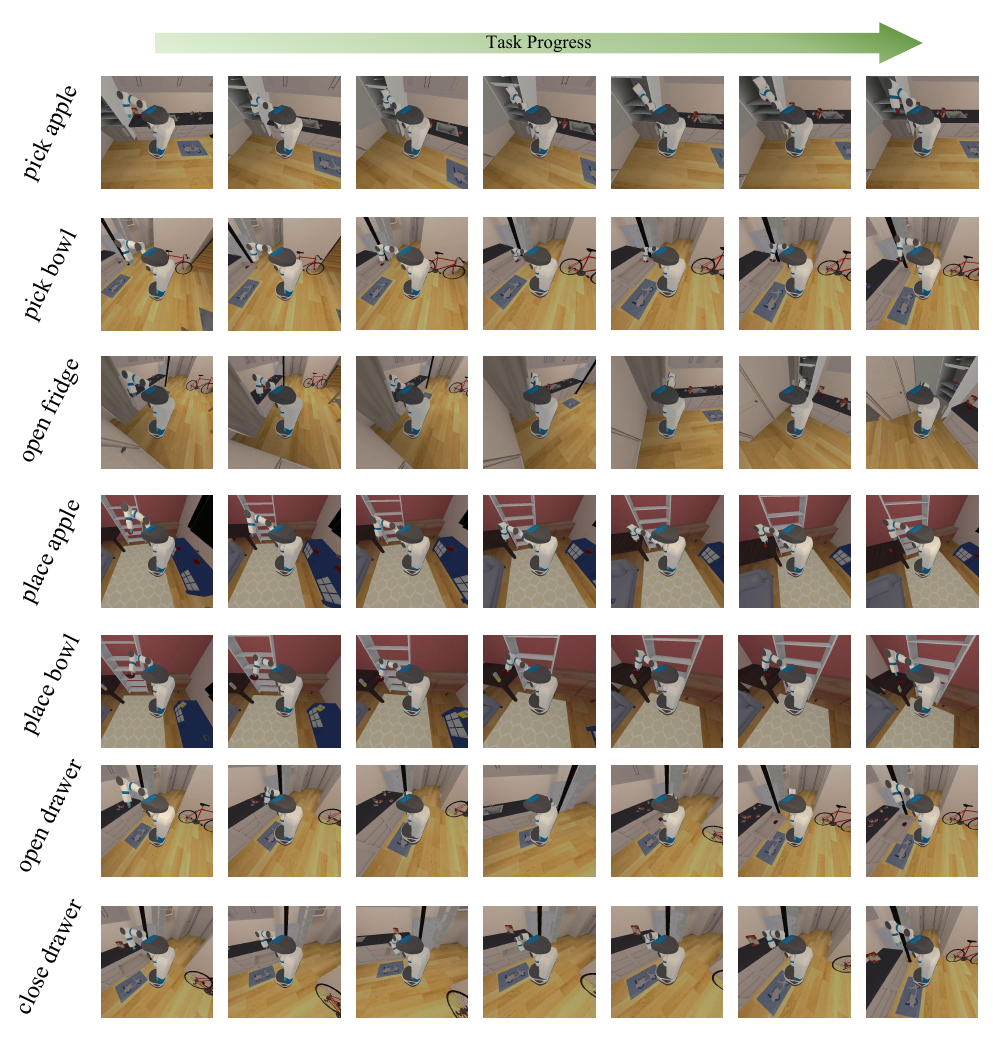}
    \captionof{figure}{Execution progress of MSHAB SetTable tasks.}
    \label{fig:set_table}
    \end{minipage}
\end{center}

\begin{center}
    \begin{minipage}{\linewidth}
    \centering
    \includegraphics[width=1.0\linewidth]{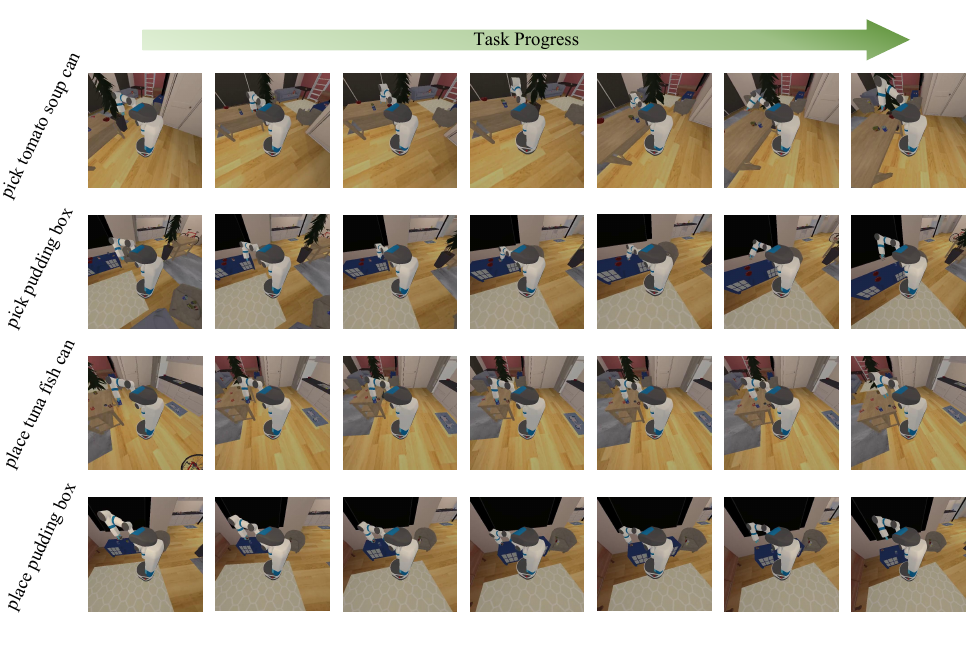}
    \captionof{figure}{Execution progress of MSHAB TidyHouse tasks.}
    \label{fig:tidy_house}
    \vspace{-1.5em}
    \end{minipage}
\end{center}

\begin{center}
    \vspace{0.5em}
    \begin{minipage}{\linewidth}
    \centering
    \includegraphics[width=1.0\linewidth]{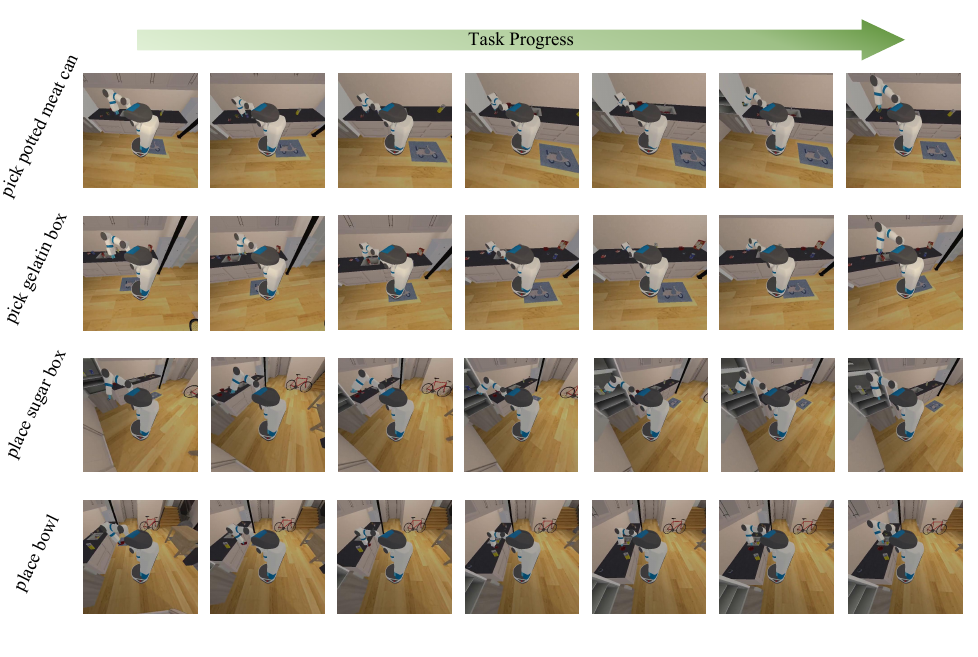}
    \captionof{figure}{Execution progress of MSHAB PrepareGroceries tasks.}
    \label{fig:prepare_groceries}
    \vspace{-0.5em}
    \end{minipage}
\end{center}

\end{document}